\def\year{2020}\relax
\documentclass[letterpaper]{article} 
\usepackage{aaai20}  
\usepackage{times}  
\usepackage{helvet} 
\usepackage{courier}  
\usepackage[hyphens]{url}  
\usepackage{graphicx} 
\usepackage{graphicx}  
\usepackage{amsmath}
\usepackage{amssymb}
\usepackage{subcaption}
\usepackage{enumitem}

\title{Graph Neural Networks for Leveraging Industrial Equipment Structure: An application to Remaining Useful Life Estimation}
\author{Jyoti Narwariya, Pankaj Malhotra, Vishnu TV, Lovekesh Vig, Gautam Shroff\\
jyoti.narwariya@tcs.com, malhotra.pankaj@tcs.com, vishnu.tv@tcs.com, lovekesh.vig@tcs.com, gautam.shroff@tcs.com\\
TCS Research, New Delhi, India
}

\begin{document}
\maketitle
\begin{abstract}
Automated equipment health monitoring from streaming multi-sensor time series data can be used to enable condition-based maintenance, avoid sudden catastrophic failures, and ensure high operational availability.
We note that most complex machinery has a well-documented and readily accessible underlying structure capturing the inter-dependencies between sub-systems or modules.
Deep learning models such as those based on recurrent neural networks (RNNs) or convolutional neural networks (CNNs) fail to explicitly leverage this potentially rich source of domain-knowledge into the learning procedure. 
In this work, we propose to capture the structure of a complex equipment in the form of a graph, and use graph neural networks (GNNs) to model multi-sensor time series data.
Using remaining useful life estimation as an application task, we evaluate the advantage of incorporating the graph structure via GNNs on the publicly available turbofan engine benchmark dataset.
We observe that the proposed GNN-based RUL estimation model compares favorably to several strong baselines from literature such as those based on RNNs and CNNs.
Additionally, we observe that the learned network is able to focus on the module (node) with impending failure through a simple attention mechanism, potentially paving the way for actionable diagnosis.
\end{abstract}

\section{Introduction}
\begin{figure}
	\centering
	\includegraphics[width=0.55\columnwidth]{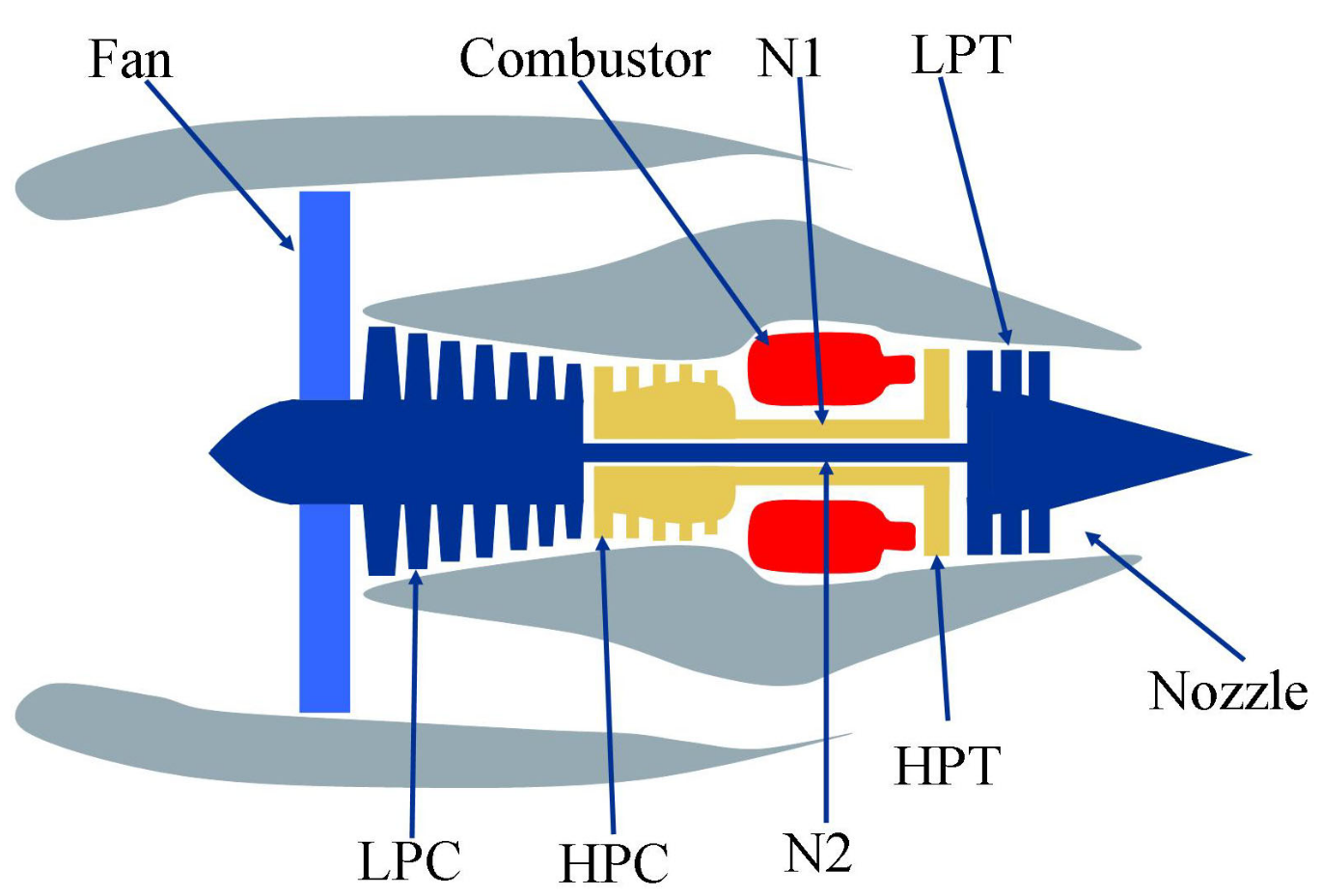}
	\includegraphics[width=0.4\columnwidth]{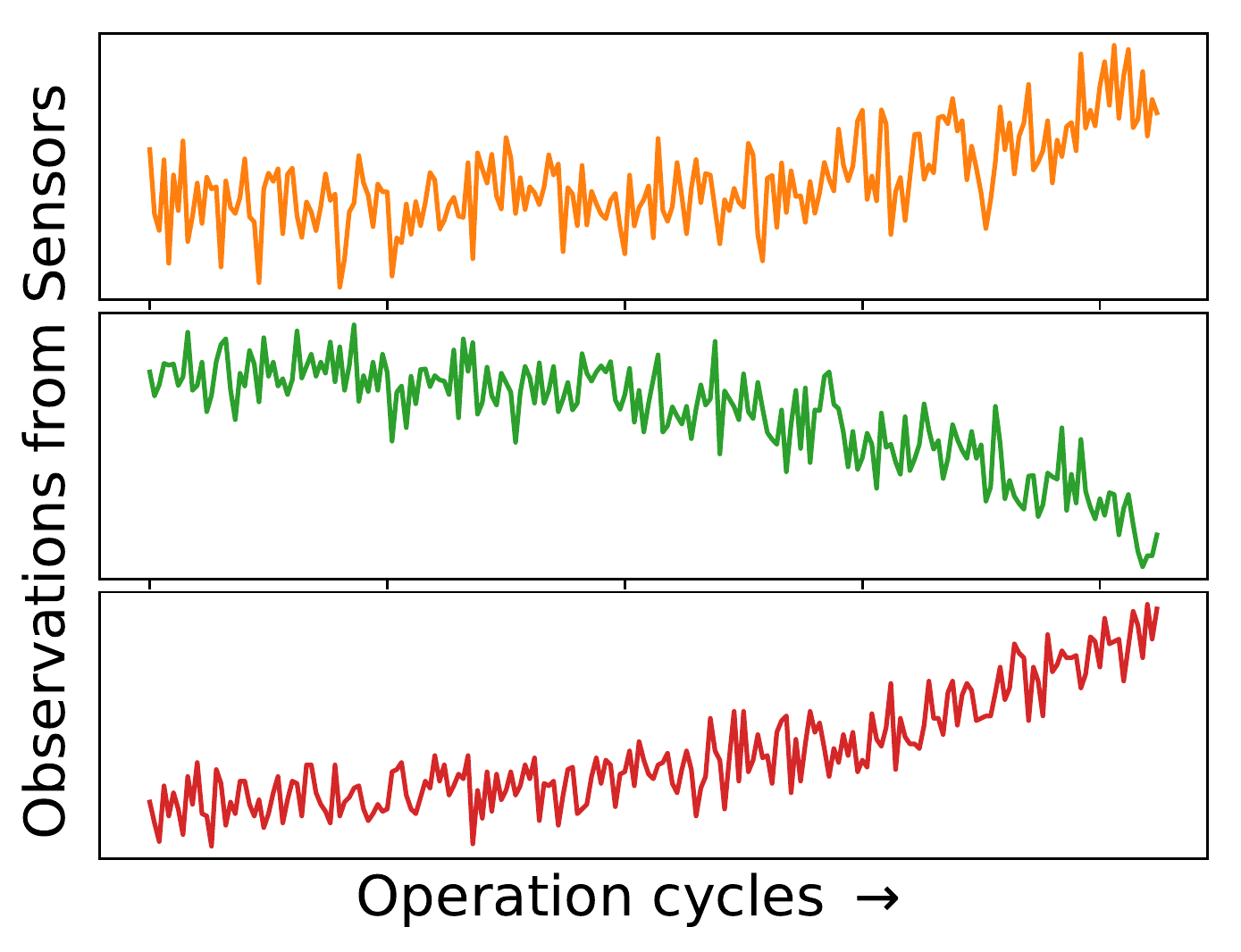}
	\caption{Left: Simplified structure of a turbofan engine depicting different modules / components [adapted from \cite{saxena2008damage}]. Right: Observations from sensors installed on an instance of the engine depicting typical degradation trends due to an impending failure. \label{fig:turbofan}}
\end{figure}

Complex industrial equipment such as engines, turbines, aircrafts, etc., are typically instrumented with a large number of sensors that result in multivariate time series data.
Most deep learning approaches model such multivariate time series data using variants of recurrent neural networks (RNNs) and convolutional neural networks (CNNs), e.g. \cite{heimes2008recurrent,p:lstm-ad,babu2016deep,zhang2016multiobjective,li2018remaining}.
These approaches often follow an ``end-to-end'' design philosophy which emphasizes minimal a priori assumptions about the system \cite{lecun2015deep}, and therefore, ignore or fail to leverage explicit structures. 
However, in most industrial setups, a complex equipment has a well-defined and documented structure: it consists of multiple interconnected modules, with the dynamics of one module affecting the dynamics of other modules. 
Fig. \ref{fig:turbofan} shows an aircraft turbofan engine with several interconnected modules.
Existing deep learning approaches fail to leverage the underlying structure of the complex equipment, and do not have the explicit capacity to reason about inter-component relations to make decisions over a structured representation of the sensor data.

Recently, a class of models for modeling and reasoning over graphs have been proposed. These include graph neural networks (GNNs) \cite{scarselli2008graph,li2015gated}, and their recent generalization in form of graph networks \cite{battaglia2018relational}. 
This class of neural networks operate on graphs, and structure their computations accordingly.
GNNs provide the desired {\em relational inductive bias} \cite{battaglia2018relational,hamrick2018relational} to solve problems with underlying structure.
For instance, NerveNet \cite{wang2018nervenet} uses Gated GNNs (GGNN) \cite{li2015gated} to learn structured policies by explicitly modeling the structure of the agent, and are better than the policies learned by models such as multi-layer perceptrons (MLPs) that simply concatenate all the observations from the environment.

In this work, we explore the applicability of GGNNs to explicitly model the underlying graph-structured mechanism of IoT-enabled complex equipment.
We represent the structure of a particular complex equipment as a directed graph where each node corresponds to a subset of sensors (e.g. those from the same module), and an edge models the relationship or dependency between two nodes or subsets of sensors (e.g. the dependence of a module on another module).
Effectively, the multivariate time series of sensor data is then represented in the graph domain to learn a GGNN model.
We use remaining useful life (RUL) estimation \cite{si2011remaining} as a target application to validate our approach.

The key contributions of this work can be summarized as follows:
\begin{itemize}
	\item We propose an approach to capture the knowledge of the structure of a complex equipment by using GGNNs. 
	To the best of our knowledge, this is the first study to evaluate the effect of introducing such relational inductive bias into the deep learning approaches for equipment health monitoring.
	\item We show the advantage of informed modularized processing of the multi-sensor time series data by grouping the sensors into meaningful subsets guided by the underlying graph structure and modules, rather than the commonly used approach that concatenates the observations from all sensors into one multivariate time series.
	\item We provide insights into the working of GGNNs for RUL estimation: we observe that the modularized processing of multivariate time series using GGNNs followed by a simple attention mechanism for aggregating information across modules can potentially allow the network to focus more on the modules with impending failures.
	
\end{itemize}

\section{Related Work\label{sec:rw}}
Recent works in \cite{sanchez2018graph,wang2018nervenet} implement an inductive bias for object- and relation-centric representations of complex dynamical systems such as a pendulum, cartpole, toy cheetah, etc.
In this work, we draw inspiration from such approaches, and show that leveraging such inductive bias can improve performance in IIoT-enabled health monitoring applications, e.g. RUL estimation. 
To the best of our knowledge, this is the first attempt to model the time series of multivariate time series data for equipment health monitoring applications while leveraging the underlying structure and semantics of the equipment.

Another line of research focuses on incorporating the semantics of the problem into the structure of deep learning models by using ontologies. 
For instance, \cite{huang2019enhancing} propose using dense layers connected as per the ontology of the manufacturing line, followed by an RNN at the top to capture the temporal dependencies.
Similarly, \cite{zhang2019modeling} attempt to model an equipment as a graph of sensor nodes: they assume a fully-connected graph where nodes correspond to sensors, and edges capture the dependencies across sensors. 
However, they do not explicitly model the dependence between various modules of an equipment. 
Our work can be seen as a generalization of these approaches as it uses the structure of equipment to guide data processing.

Several variants of deep neural networks including CNNs and RNNs have been proposed for equipment health monitoring and RUL estimation, e.g. \cite{heimes2008recurrent,p:lstm-ad,babu2016deep,li2018remaining,gugulothu2017pred}.
However, most of these approaches consider a flat concatenation of readings or observations from all the sensors as a multivari￼
ate input to the neural network, and ignore the structure of the underlying system or mechanism from which the data is generated. 
In this work, we show that grouping the sensors into meaningful inter-dependent subsets and processing them separately before the final concatenation step yields superior performance. 

\section{Problem Setup\label{sec:overview}}

\begin{figure*}[h]
	\centering
	\begin{subfigure}{0.7\textwidth}
		\includegraphics[width=\textwidth,trim={2.3cm 2cm 1.5cm 1cm}, clip]{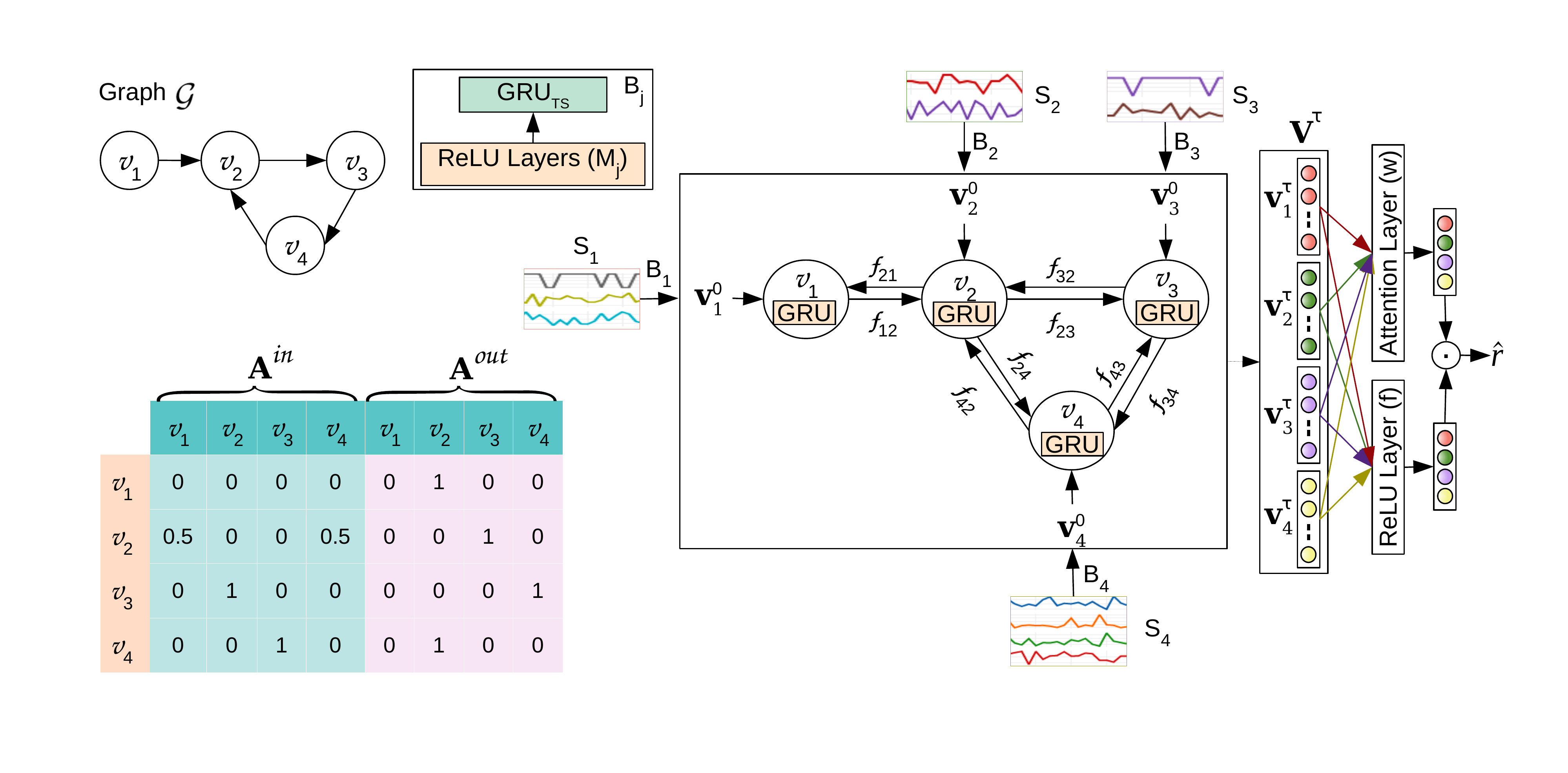}
		\caption{GGNN-based Metric Regression (GNMR) \label{fig:GGNN}}
	\end{subfigure}
~~~~~~~~~~
	\begin{subfigure}{0.15\textwidth}
		\centering
		\includegraphics[width=\textwidth,trim={0cm 2cm 1cm 1cm}, clip]{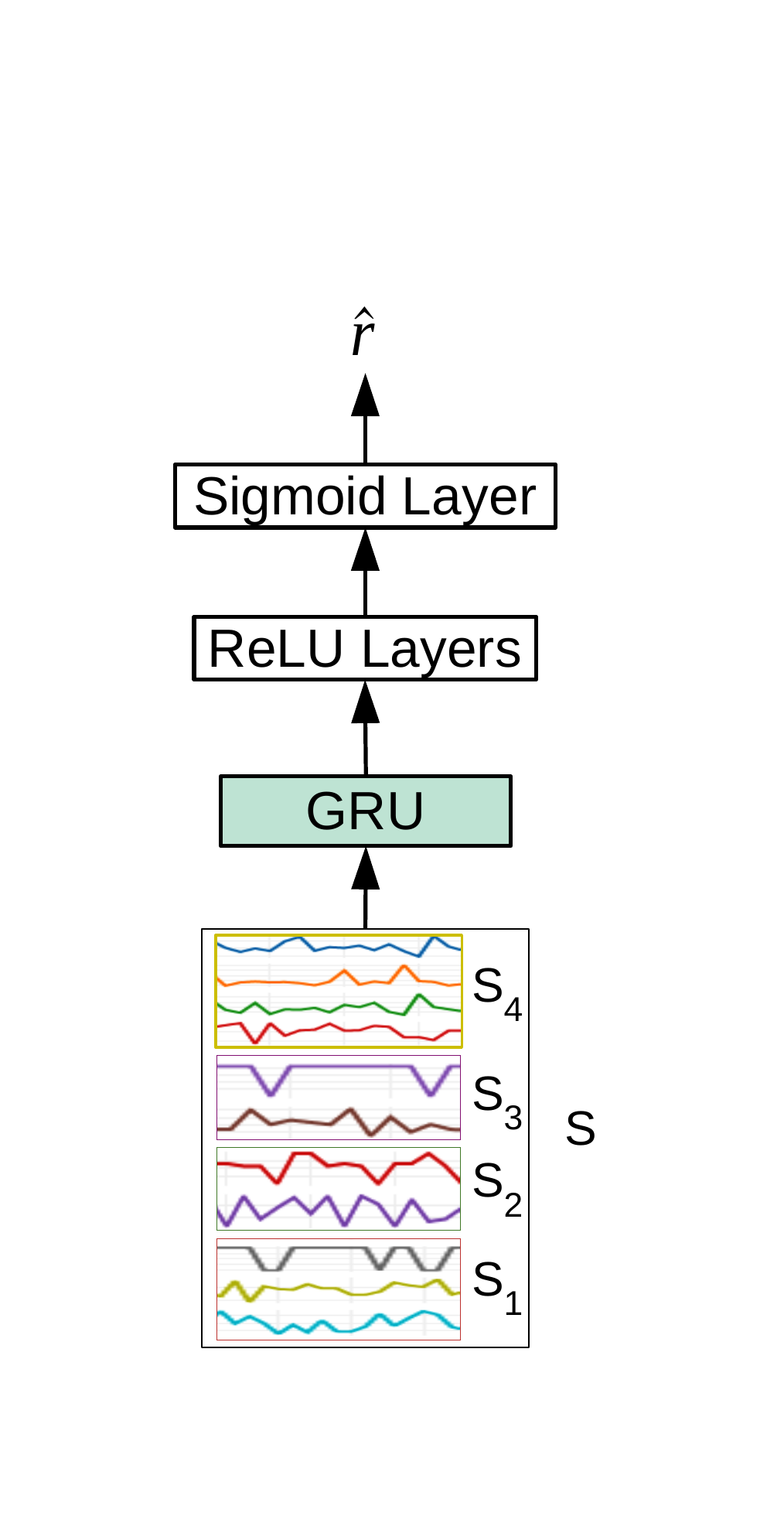}
		\caption{GRU-based Metric Regression (GRU-MR) \label{fig:GRU-MR}}
	\end{subfigure}
	
	\caption{Illustrative process flow for the proposed GGNN-based Metric Regression (GNMR) and the traditional GRU-based Metric Regression. While GRU-MR ingests observations from all sensors at once, GNMR processes observations grouped based on the equipment structure. $f_{ij}$ refers to one or more feedforward layers with leaky ReLUs that capture the influence of node $j$ on node $i$. $\mathbf{v}_j^0$ denotes the initial representation of node $j$ and $\mathbf{v}_j^\tau$ denotes the final representation of node $j$ after $\tau$ propagation steps. (Best viewed in the electronic version by zooming-in.) \label{fig:flowchart}}
\end{figure*}

We consider the scenario where a complex equipment consists of multiple modules (sub-systems) connected to each other in a known fashion. 
Let $\mathcal{S}$ denote the set of all the sensors installed to monitor various parameters across various modules of the equipment.
The dynamical behavior of any module is observed via the multivariate time series corresponding to a fixed and known subset of sensors (parameters) associated with that module.
We represent the equipment as a directed graph $\mathcal{G} = (\mathcal{V}, \mathcal{E})$, where $v_j\in \mathcal{V}$ (for $j=1\ldots |\mathcal{V}|$) is a node in the graph that corresponds to a subset of sensors $\mathcal{S}_j \subset \mathcal{S}$ associated with the module indexed by $j$, $e_{jk} = (v_j,v_k) \in \mathcal{E}$ is a directed edge from node $v_j$ to $v_k$ that models the influence of $\mathcal{S}_k$ on $\mathcal{S}_j$.
Note that, in general, $\mathcal{S}_j \cap \mathcal{S}_k \neq \emptyset$, such that a sensor can be associated with more than one node: for example, a sensor measuring the ambient temperature can be associated with all nodes.
We consider the supervised learning setting: we are given a fixed graph structure $\mathcal{G}$ and a learning set $\mathcal{D} = \{\mathbf{x}^{i}_1 \ldots \mathbf{x}^{i}_{|\mathcal{V}|},r^{i}\}^n_{i=1}$ of $n$ time series, where $r^i \in \mathbb{R}$ is the target value, and $\mathbf{x}^{i}_1 \ldots \mathbf{x}^{i}_{|\mathcal{V}|}$ denotes the $|\mathcal{V}|$ multivariate time series associated with the $|\mathcal{V}|$ nodes of $\mathcal{G}$.
Here, $\mathbf{x}_j^i= \mathbf{x}_{j,1}^i \ldots \mathbf{x}_{j,T}^i$ denotes the $p_j$-dimensional multivariate time series corresponding to node $v_j$ for time $t=1\ldots T$, where $\mathbf{x}_{j,t}^i \in \mathbb{R}^{p_j}$ and $p_j = |\mathcal{S}_j|$ denote the number of sensors in $\mathcal{S}_j$.

For the RUL estimation task, the $n$ time series are collected from one or more instances (installations) of an equipment with structure $\mathcal{G}$, and the target variable $r^i \in \mathbb{R}$ corresponds to the RUL value at time $T^i$, e.g. in terms of remaining cycles of operation or remaining operational hours.
RUL estimation is then a metric regression task where the goal is to map $\mathbf{x}^{i}_1 \ldots \mathbf{x}^{i}_{|\mathcal{V}|}$ to $r^i$.
Let $F^i$ denote the total operational life of an instance $i$ till the failure point, s.t. at any time $T^i \leq F^i$, the target RUL is given by $r^{i}=F^i-T^i$.
Furthermore, as in a typical practical setting, we assume that all instances of the equipment have the same underlying graph structure $\mathcal{G}$, i.e. the different modules of the equipment are connected to each other in same fashion. 

\section{Approach\label{sec:approach}}
As illustrated in Fig. \ref{fig:GGNN}, each multivariate time series $\mathbf{x}_j$ is processed by a neural network $B_j$ (for $j=1,\ldots, |\mathcal{V}|$) to obtain a fixed-dimensional initial node representation vector $\mathbf{v}_j^0 \in \mathbb{R}^d$. 
This initial representation $\mathbf{v}_j^0$ is then updated using the representations of neighboring nodes defined by $\mathcal{G}$ using a message passing mechanism to obtain $\mathbf{v}_j^\tau$.
Finally, an attention mechanism is used to combine the final node representations to obtain an RUL estimate $\hat{r}$ for the equipment. 
We refer to this approach as \textbf{GNMR} (Gated \textbf{G}raph Neural \textbf{N}etworks for \textbf{M}etric \textbf{R}egression). The entire computation flow of GNMR is differentiable end-to-end and the associated parameters are learned via stochastic gradient descent.
Next, we describe these steps in more detail.

\subsection{Learning Node Representations from Time Series}
The $p_j$-dimensional time series $\mathbf{x}_j$ at node $v_j$ is processed by $B_j$ to obtain $\mathbf{v}_j^0$. 
We consider a gated recurrent units (GRU)-based RNN \cite{cho2014learning} for processing this time series. 
In general, $p_j$ is different across the nodes implying that we need to learn and maintain $|\mathcal{V}|$ GRU networks.
This can pose scalability issues for graphs with large number of nodes. 
It is, therefore, desirable to use a common GRU, which we refer to as $GRU_{TS}$, to process the multivariate time series from all the nodes so as to keep the number of trainable parameters of the network within manageable limits. 
To this end, any point $\mathbf{x}_{j,t} \in \mathbb{R}^{p_j}$ ($t=1\ldots T$) at node $v_j$ is first processed via a node-specific feedforward network $M_j$ to obtain a fixed $d$-dimensional vector $\tilde{\mathbf{x}}_{j,t} \in \mathbb{R}^{d}$. 
Note that $d$ is same across nodes allowing us to use the common $GRU_{TS}$ for further processing of the resulting time series $\tilde{\mathbf{x}}_{j}$ to obtain the initial representation $\mathbf{v}_{j}^0$.
Effectively, $B_j$ consisting of $M_j$ and $GRU_{TS}$ maps the input time series $\mathbf{x}_j$ to $\mathbf{v}_{j}^0$.

\subsection{Message Passing Across Nodes}
While the node-level representation $\mathbf{v}_{j}^0$ obtained from $B_j$ can capture the dependencies across sensors in $\mathcal{S}_j$, it ignores the dependencies across nodes. 
It is desirable to leverage the representations of neighboring nodes to capture the dependencies between nodes, and then aggregate them to obtain a representation for the overall dynamics of the system.
To achieve this, the representations for each node are iteratively updated by a GGNN using the representations of the neighboring nodes as described next.

Consider two normalized adjacency matrices $\mathbf{A}^{in} \in \mathbb{R}^{|\mathcal{V}|\times \mathcal{V}|}$ and $\mathbf{A}^{out} \in \mathbb{R}^{|\mathcal{V}|\times \mathcal{V}|}$ corresponding to the incoming and outgoing edges in graph $\mathcal{G}$, as illustrated in Fig. \ref{fig:GGNN}.
Let $\mathbf{v}_{j}^m \in \mathbb{R}^d$ correspond to the $j$th row of matrix $\mathbf{V}^m \in \mathbb{R}^{|\mathcal{V}|\times d}$, and denote the representation (or embedding) for node $v_j$ after $m$ message propagation steps.
GGNN takes $\mathbf{A}^{in}$, $\mathbf{A}^{out}$, and the initial node representations $\mathbf{V}^0$ as input, and returns an updated set of representations $\mathbf{V}^\tau$ after $\tau$ iterations of message propagation across nodes in the graph s.t. $[\mathbf{v}^\tau_{1},\mathbf{v}^\tau_{2},\ldots,\mathbf{v}^\tau_{|\mathcal{V}|}] = G(\mathbf{A}^{in},\mathbf{A}^{out},\mathbf{V}^0;\boldsymbol{\theta}_g)$, where $\boldsymbol{\theta}_g$ represents the parameters of the GGNN function $G$.
For any node $v_j$ in the graph and message propagation step $m$, the previous representation of the node $\mathbf{v}_j^{m-1}$, and the aggregated representation $\mathbf{a}_j^{m}$ of its neighboring nodes (as obtained via Eqs. \ref{eq1}-\ref{eq3} below) are used to iteratively update the representation of the node $\tau$ times.

More specifically, the representation of node $v_j$ in the message propagation step $m~(=1 \ldots \tau)$ is updated as follows:
\begin{align}
\mathbf{p}_{ij}^m &= f_{ij}(\mathbf{v}_i^{m-1};\boldsymbol{\theta}_{ij}), ~\mathbf{p}_{jk}^m = f_{jk}(\mathbf{v}_k^{m-1};\boldsymbol{\theta}_{jk}) \label{eq1},\\
\mathbf{P}_{1j}^m &= [\mathbf{p}_{1j}^m \ldots \mathbf{p}_{|\mathcal{V}|j}^m]^\top, ~\mathbf{P}_{2j}^m = [\mathbf{p}_{j1}^m \ldots \mathbf{p}_{j|\mathcal{V}|}^m]^\top \label{eq2},\\
\mathcal{\mathbf{a}}_{j}^m &= [\mathcal{\mathbf{A}}^{in}_{j:} \mathbf{P}_{1j}^m, \mathcal{\mathbf{A}}^{out}_{j:} \mathbf{P}_{2j}^m]^\top \label{eq3},\\
\mathcal{\mathbf{z}}_{j}^m &= \sigma (\mathbf{W}_z \mathcal{\mathbf{a}}_{j}^m + \mathbf{U}_z\mathbf{v}_{j}^{m-1}),\label{eq4}\\
\mathbf{r}_{j}^m &= \sigma (\mathbf{W}_r \mathcal{\mathbf{a}}_{j}^m + \mathbf{U}_r\mathbf{v}_{j}^{m-1})\label{eq5},\\
\hat{\mathbf{v}}_{j}^m &= \tanh (\mathbf{W}_o \mathcal{\mathbf{a}}_{j}^m + \mathbf{U}_o (\mathbf{r}_{j}^m \odot \mathbf{v}_{j}^{m-1}))\label{eq6},\\
\mathbf{v}_{j}^m &= (1-\mathbf{z}_{j}^m)\odot \mathbf{v}_{j}^{m-1} + \mathbf{z}_{j}^m \odot \hat{\mathbf{v}}_{j}^m \label{eq7},
\end{align}
where $i,k=1,\ldots,|\mathcal{V}|$, $\mathbf{A}^{in}_{j:}$ and $\mathbf{A}^{out}_{j:}$ denote the $j$-th row of $\mathbf{A}^{in}$ and $\mathbf{A}^{out}$, respectively. 
Here, $\mathbf{A}^{in}$ and $\mathbf{A}^{out}$ allow to capture the information from upstream and downstream nodes in the system, respectively.
$f_{ij}$ denotes feedforward ReLU layer(s) with parameters $\boldsymbol{\theta}_{ij}$ that computes the contribution (message) from $v_i$ to $v_j$ if there is an incoming edge from $v_i$ to $v_j$, i.e. when $e_{ij}\in \mathcal{E}$.
Then, $\mathbf{p}_{ij}^m \in \mathbb{R}^d$ denotes the message from $v_i$ to $v_j$ corresponding to edge $e_{ij}$. 
Similarly, $f_{jk}$ computes the message from $v_k$ to $v_j$ if there is an outgoing edge from $v_j$ to $v_k$, i.e. when $e_{jk}\in \mathcal{E}$.
$\mathbf{P}_{1j}^m$ and $\mathbf{P}_{2j}^m \in \mathbb{R}^{|\mathcal{V}|\times d}$ denote the matrices that contain the information from the incoming and outgoing edges with $v_j$ as starting and ending node, respectively.
For $e_{ij} \notin \mathcal{E}$, $f_{ij}$ simply returns $\mathbf{0} \in \mathbb{R}^d$.
The trainable parameters $\boldsymbol{\theta}_{ij}$, $\boldsymbol{\theta}_{ji}$, $\mathbf{W}_{(.)}$ and $\mathbf{U}_{(.)}$ of appropriate dimensions constitute $\boldsymbol{\theta}_g$, $\sigma (.)$ is the sigmoid function, and $\odot$ is the element-wise multiplication operator.
Eqs. \ref{eq4}-\ref{eq7} are the computations equivalent to a gated recurrent unit (GRU) network.

Note that for large graphs with many nodes and edges, the number of functions $f_{ij}$ and their associated parameters $\boldsymbol{\theta}_{ij}$ can be large. However, in many practical applications such as a power plant or a water treatment plant \cite{goh2016dataset}, nodes typically have an associated type with more than one node belonging to the same type, e.g. multiple water tanks in a water treatment plant. 
Under such scenarios, it may be suitable to tie the parameters of the edge functions for a given pair of node types.

\subsection{Attention Mechanism to Aggregate Node-level Representations}
The final representations $\mathbf{v}_1^{\tau},\ldots, \mathbf{v}_{|\mathcal{V}|}^{\tau}$ can be aggregated to get a graph-level output (RUL estimate in our case) by using an attention mechanism, e.g. as used in \cite{li2015gated}.
In this work, we consider a simple variant of this attention mechanism: 
For each node, we use the concatenated vector $\tilde{\mathbf{v}}_j^\tau = [\mathbf{v}_j^0,\mathbf{v}_j^\tau,T,node\_type_j]$ as inputs to two parallel feedforward layers $f_1$ and $f_2$ to obtain $f_1(\tilde{\mathbf{v}}_j^\tau) \in \mathbb{R}$ and $\hat{r}_j = f_2(\tilde{\mathbf{v}}_j^\tau) \in \mathbb{R}$. 
Here, $node\_type_j$ is a one-hot vector of length $|\mathcal{V}|$, and is set to 1 for $j$th position, and 0 otherwise. 
Also, $T$ is used as an additional input for the RUL estimation task as the total life passed can be a useful feature to estimate the wear-and-tear of the system.
We apply softmax over the values from $f_1$ to obtain attention weight $w_j = \frac{\textnormal{exp}(f_1(\tilde{\mathbf{v}}_j^\tau)}{\sum_i\textnormal{exp}(f_1(\tilde{\mathbf{v}}_i^\tau))}$ for node $v_j$.
The final RUL estimate is then given by 
\begin{equation}
\hat{r} = \sum_{j=1}^{\mathcal{|V|}}w_j \hat{r}_j.\label{eq:attn}
\end{equation}
This can be interpreted as assigning a weightage $0\leq w_j\leq 1$ to the node $v_j$ while $\hat{r}_j = f_2(\tilde{\mathbf{v}}_j^\tau)$ denotes the RUL estimate as per node $v_j$.

Given the training set $\mathcal{D}$, the loss function used for training is then given by $\mathcal{L}_\mathcal{D} = \frac{1}{n}\sum_i^{n} (r^i-\hat{r}^{i})^2$,
with learning parameters being $\boldsymbol{\theta}_g$, parameters of $GRU_{TS}$, and the parameters of the feedforward layers ($M_i$s, $f_1$, and $f_2$) with leaky ReLU units.

\section{Experimental Evaluation\label{sec:exp}}
\begin{figure}
	\centering
	\includegraphics[width=0.5\linewidth,trim={1.3cm 1cm 5.9cm 1cm},clip]{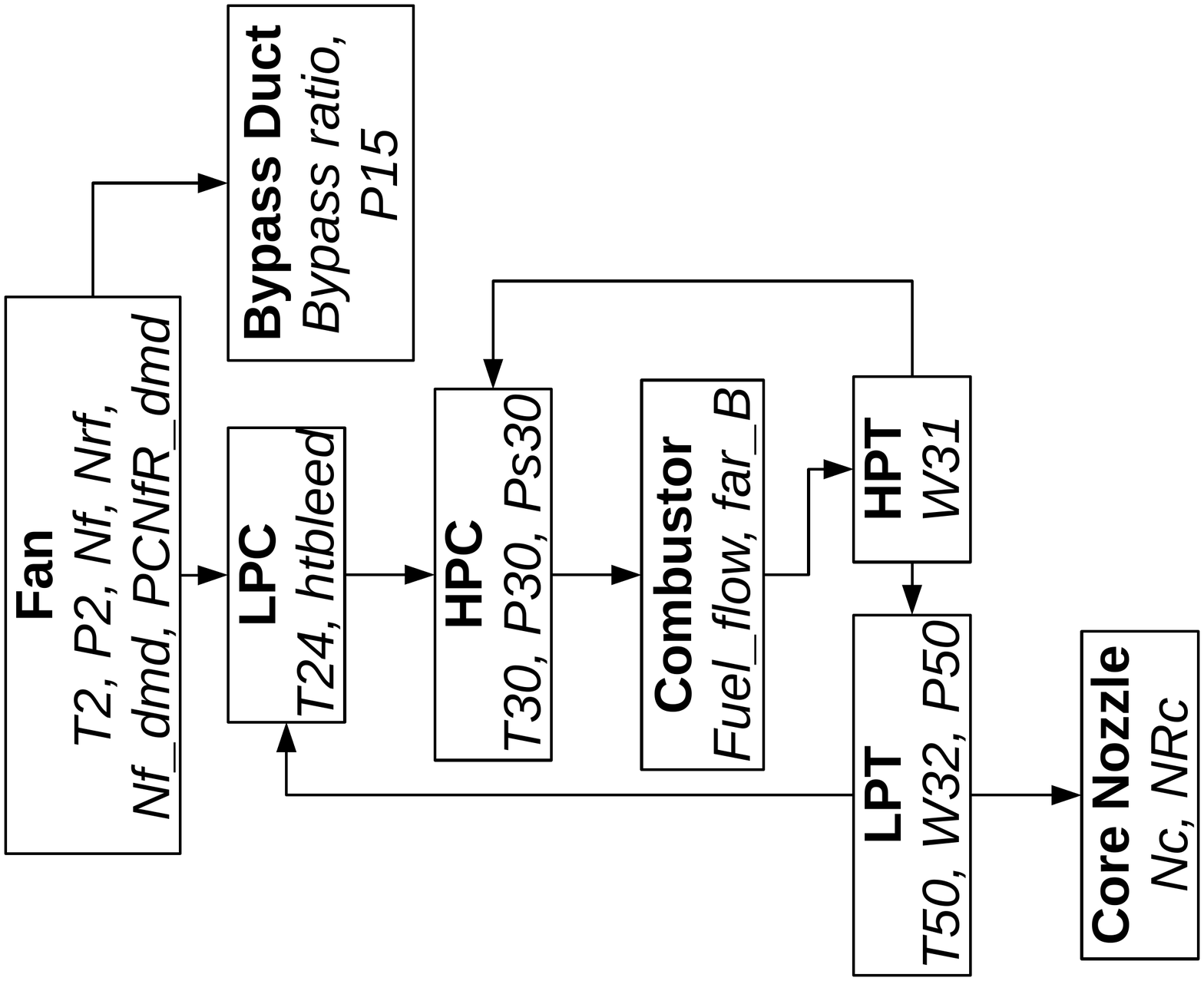}
	\caption{Original graph structure used for the experiments. Each node corresponds to a module (in bold) with an associated subset of sensors (in italics). Refer Table 2 in \cite{saxena2008damage} for details of sensor names. \label{fig:structure-used}}
\end{figure}
We investigate if \textbf{GNMR} is able to leverage the graph structure to improve upon other strong baselines from literature that consider a simple concatenation of observations as a multivariate input: CNNs \cite{babu2016deep,li2018remaining}, LSTMs \cite{zheng2017long}, and Deep Belief Networks \cite{zhang2016multiobjective}. We additionally implement \textbf{GRU-MR}: a GRU-based RUL estimation model on lines of Metric Regression approach proposed in \cite{zheng2017long} as depicted in Fig. \ref{fig:GRU-MR}.
We ensure comparable hyperparameter settings as well as the same train, validation and test splits for GRU-MR and the proposed GNMR, as detailed later.
We further study the sensitivity of the approach to the knowledge of the graph structure by synthetically combining or segregating nodes (modules) in the original graph structure.
To study other ways of combining time series of sensors, we also consider reducing the number of input dimensions for GRU via principal components analysis within the GRU-MR framework, and refer to it as \textbf{PCA-GRU-MR}. We report results for 5 PCA components (capturing 85\% variance in the data), i.e. 5-dimensional input to GRU-MR in Table \ref{tab:litComp}.

\textbf{Datasets}: We use the four publicly available aircraft turbofan engine benchmark datasets\footnote{\url{https://ti.arc.nasa.gov/tech/dash/groups/pcoe/prognostic-data-repository/#turbofan}} FD001-FD004, as introduced in \cite{saxena2008damage}. 
We use the equipment structure information as depicted in Figs. \ref{fig:turbofan} and \ref{fig:structure-used} (refer \cite{saxena2008damage} for details). 
Each dataset contains a pre-defined train-test split. 
We further use a random 80-20 split of the original train split to obtain a train and a validation set.
The hold-out validation set is used for hyperparameter tuning.

\begin{table*}
	\centering
	\caption{Performance comparison in terms of RMSE and S. Best performance is shown in bold and second-best is underlined. \label{tab:litComp}}
	\scalebox{0.8}{
		\begin{tabular}{|c|cc|cc|cc|cc|cc|}
			\hline
			\multicolumn{1}{|c|}{Dataset $\rightarrow$} & \multicolumn{2}{c}{\bfseries FD001} & \multicolumn{2}{|c|}{\bfseries FD002} & \multicolumn{2}{c}{\bfseries FD003} & \multicolumn{2}{|c|}{\bfseries FD004} & \multicolumn{2}{|c|}{\bfseries Average Rank}\\ \hline
			Method $\downarrow$ & RMSE & S & RMSE & S & RMSE & S & RMSE & S & RMSE &S\\
			\hline
			CNN-MR \cite{babu2016deep} & 18.45 &1,287 &30.29 &13,570&19.82&1596&29.16 &7,886 &8&7.75 \\ 
			LSTM-MR \cite{zheng2017long} & 16.14 &338 &24.49&4,450&16.18&852&28.17 &5,550 &6.5&5.25\\ 
			MODBNE \cite{zhang2016multiobjective} & 15.04 &334&25.05&5,585&\textbf{12.51}&422&28.66 &6,558&4.75&5.25 \\
			CNN + FNN \cite{li2018remaining} & \underline{12.61} &\underline{274} &22.36&10,412&12.64&\textbf{284}&23.31 &12,466 &3.25&4.5\\
			\hline
			GRU-MR (ours) &15.36&481&22.43&{3,391}&\underline{12.52}&\underline{339}&{22.96}&{2,964} &3.75&3.75\\
			
            PCA-GRU-MR (ours) &15.60&469&22.92&{3,916}&{13.37}&860&{22.41}&\textbf{2,637} &5&4.5\\
			GNMR (proposed, $\tau=0$) &12.73&302&\underline{21.38}&\textbf{3,148}&{13.06}&{366}&\underline{21.81}&3,414&\underline{2.75}&\underline{2.75} \\
			GNMR (proposed) &\textbf{12.14}&\textbf{212}&\textbf{20.85}&\underline{3,196}&13.23&370&\textbf{21.34}&\underline{2,795}&\textbf{2}&\textbf{2.25} \\
			\hline
	\end{tabular}}
	
\end{table*}

\begin{table*}
	\centering
	\caption{Effect of varying the graph structure. \label{tab:absComp}}
	\scalebox{0.8}{
		\begin{tabular}{|c|c|cc|cc|cc|cc|cc|}
			\hline
			\multicolumn{2}{|c|}{Dataset $\rightarrow$} & \multicolumn{2}{c}{\bfseries FD001} & \multicolumn{2}{|c|}{\bfseries FD002} & \multicolumn{2}{c}{\bfseries FD003} & \multicolumn{2}{|c|}{\bfseries FD004} & \multicolumn{2}{|c|}{\bfseries Average Rank}\\ \hline
			Method $\downarrow$& $\mathcal{|V|}$ &RMSE & S & RMSE & S & RMSE & S & RMSE & S & RMSE & S\\
			\hline
			{Original Graph} &8&\textbf{12.14}&\textbf{212}&\textbf{20.85}&\textbf{3,196}&\textbf{13.23}&\textbf{370}&\underline{21.34}&2,795&\textbf{1.25}&\textbf{1.5} \\
			One node for all sensors &1&13.20&321&22.36&4,072&13.65&\underline{378}&22.04&\underline{2,747}&4.25&3 \\
			Reduced Nodes &4&12.41&299&22.85&4,798&15.26&981&21.87&3,034&4.25&4.5 \\			
			Increased Nodes &13&\underline{12.15}&\underline{214}&22.20&4,512&13.63&623&\textbf{20.95}&3,813&\underline{2.25}&3.75 \\
			One node per sensor &21&13.86&293&\underline{21.99}&\underline{3,317}&\underline{13.39}&431&21.48&\textbf{2,562}&3&\underline{2.25} \\
			\hline
	\end{tabular}}

\end{table*}

\textbf{Performance Metrics}: We use the standard metrics RMSE and Timeliness Score (S) as introduced in \cite{saxena2008damage}. 
Let $e^i$ = $\hat{r}^{i}-r^i$ denote the error in RUL estimate for $i$th test instance, 
then $RMSE = \sqrt{\frac{1}{n}\sum_{i=1}^n (e^i)^2}$ and $S=\sum_{i=1}^n(\textnormal{exp}(\gamma . |e^i|)-1)$, $n$ is the number of test instances, $\gamma=1/u_1$ if $e^i < 0$, else $\gamma=1/u_2$. 
Usually, $u_1 > u_2$ such that late predictions are penalized more compared to early predictions. 
We consider $u_1=13$ and $u_2=10$ as used in all the baselines. 
Lower values of RMSE and $S$ indicate better performance.

\textbf{Hyperparameters}: We use batch size of 32, and a dropout rate of 0.2 for all feedforward (leaky ReLU) layers.
We use 2 leaky ReLU layers for each $M_j$ and $f_{ij}$ in GNMR as well as for the pre-final leaky ReLU layers of GRU-MR (refer Fig. \ref{fig:flowchart}).
We use Adam optimizer with initial learning rate of 0.001 which is reduced every 10 epochs by a factor of $1/\sqrt{2}$.
The number of hidden units, same as $d$, for all feedforward and recurrent layers is chosen from \{30,60\}.
We use message propagation steps $\tau = \{0,2,4\}$ for GGNN. The number of hidden layers for $GRU_{TS}$ in GNMR and for GRU-MR is chosen from \{2,3,4\}. 
All hyperparameters are selected via grid search based on validation RMSE.

\subsection{Results and Observations}
	\textbf{(1)} \textit{Comparison with baselines}: From Table \ref{tab:litComp}, we observe that GGNN performs better than GRU-MR, PCA-GRU-MR as well as other baselines across most datasets. 
	GNMR has the highest average rank of 2.0 based on both RMSE and S. 
	Furthermore, the special case of GNMR with no message propagation across nodes, i.e. $\tau=0$, also compares favorably to GRU-MR and other baselines. 
	These results suggest that meaningful grouping of the sensors into nodes based on the knowledge of the modular graph structure of the equipment is advantageous over methods that consider a concatenated vector of all the sensors as one input. 
	Also, allowing for message propagation across nodes gives further improvement in results indicating the advantage of modeling the dependencies between modules.
	
	\noindent \textbf{(2)} \textit{Effect of varying the graph structure}: We consider two scenarios to study the sensitivity of results to the exact knowledge of graph structure: for the \textit{Increased Nodes} scenario, any original node $v_j$ in $\mathcal{G}$ with $|\mathcal{S}_j|>1$ is split into two nodes, say $v_j^1$ and $v_j^2$, by randomly distributing the sensors in $\mathcal{S}_j$ to the two nodes such that each new node gets half the sensors.
	Furthermore, the nodes $v_j^1$ and $v_j^2$ thus created are connected to each other.
	Also, if $e_{jk}\in \mathcal{E}$, then both $v_j^1$ and $v_j^2$ are additionally connected to the new nodes $v_k^1$ and $v_k^2$ obtained from $v_k$.
	For the \textit{Reduced Nodes} scenario, when combining two neighboring nodes $v_j$ and $v_k$ into a new node, say $v_{jk}$, we have $\mathcal{S}_{jk} = \mathcal{S}_j \cup \mathcal{S}_k$, and an edge exists between any two new nodes if there was an edge between the original nodes from which the new nodes were created.
	We also consider the limiting cases of \textit{One node per sensor} and \textit{One node for all sensors} for the \textit{Increased Nodes} and \textit{Reduced Nodes} scenarios, respectively.
	
	From Table \ref{tab:absComp}, we observe that the performance degrades as we reduce the nodes in the graph, with \textit{Reduced Nodes} and \textit{One node for all sensors} being the worst models.
	On the other hand, the graph with \textit{Increased Nodes} performs the same as the \textit{Original Graph} for FD001 while being better on FD004. Nevertheless, the \textit{Original Graph} still gives best performance on average across datasets. 
	\textit{Increased Nodes} and \textit{One node per sensor} have the second-best performance. However, it is to be noted that the models with increased nodes have a much larger number of MLPs for message propagation (refer Eqs. \ref{eq1}-\ref{eq2}) due to increased number of edges, making it computationally more expensive when compared to the original graph.

\begin{figure}[th]
	\centering
	\begin{subfigure}{0.42\columnwidth}
		\centering
		\includegraphics[width=\linewidth,height=0.8\linewidth]{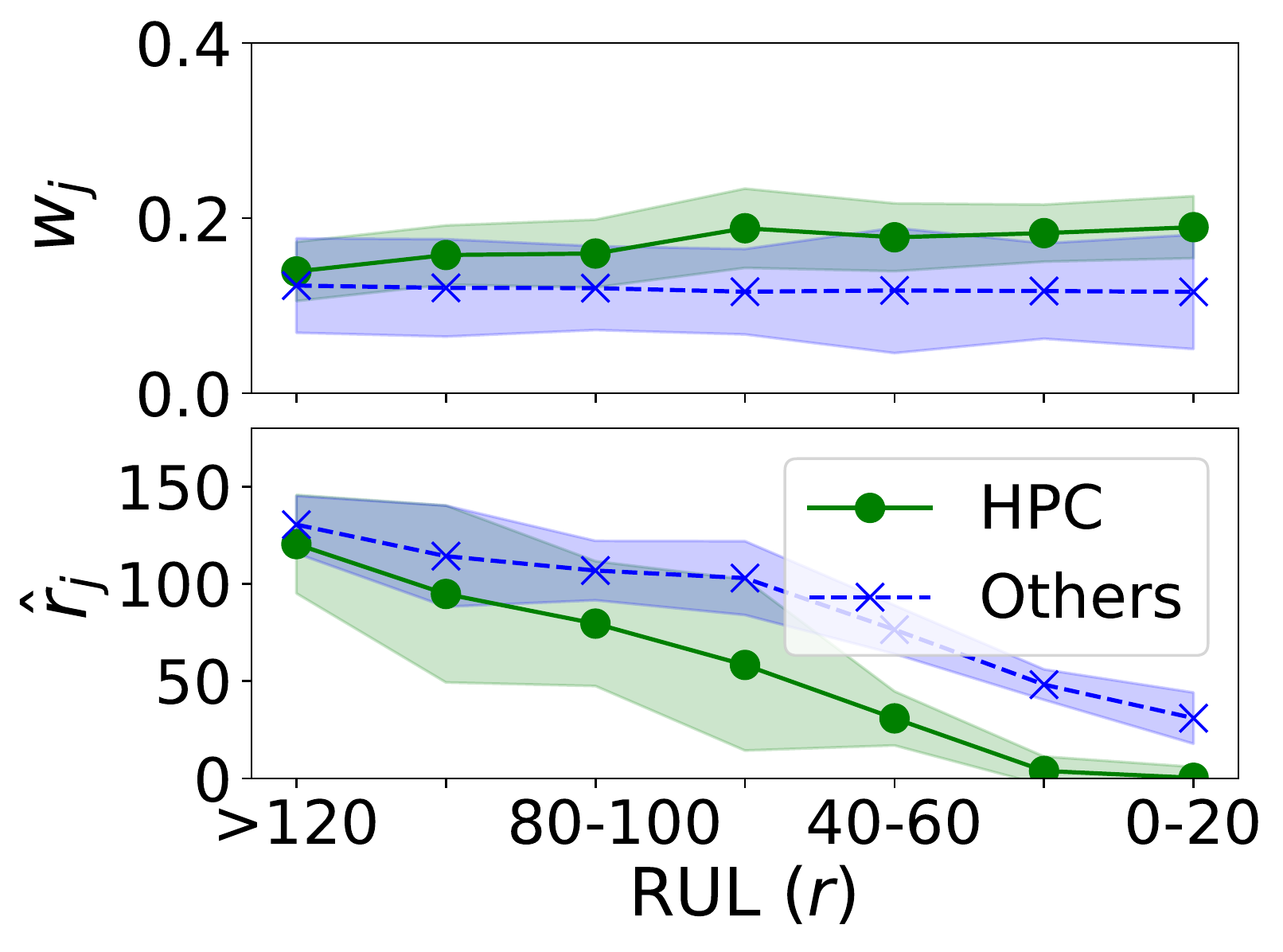}
		\caption{FD001}
	\end{subfigure}
	~~~
	\begin{subfigure}{0.42\columnwidth}
		\centering
		\includegraphics[width=\linewidth,height=0.8\linewidth]{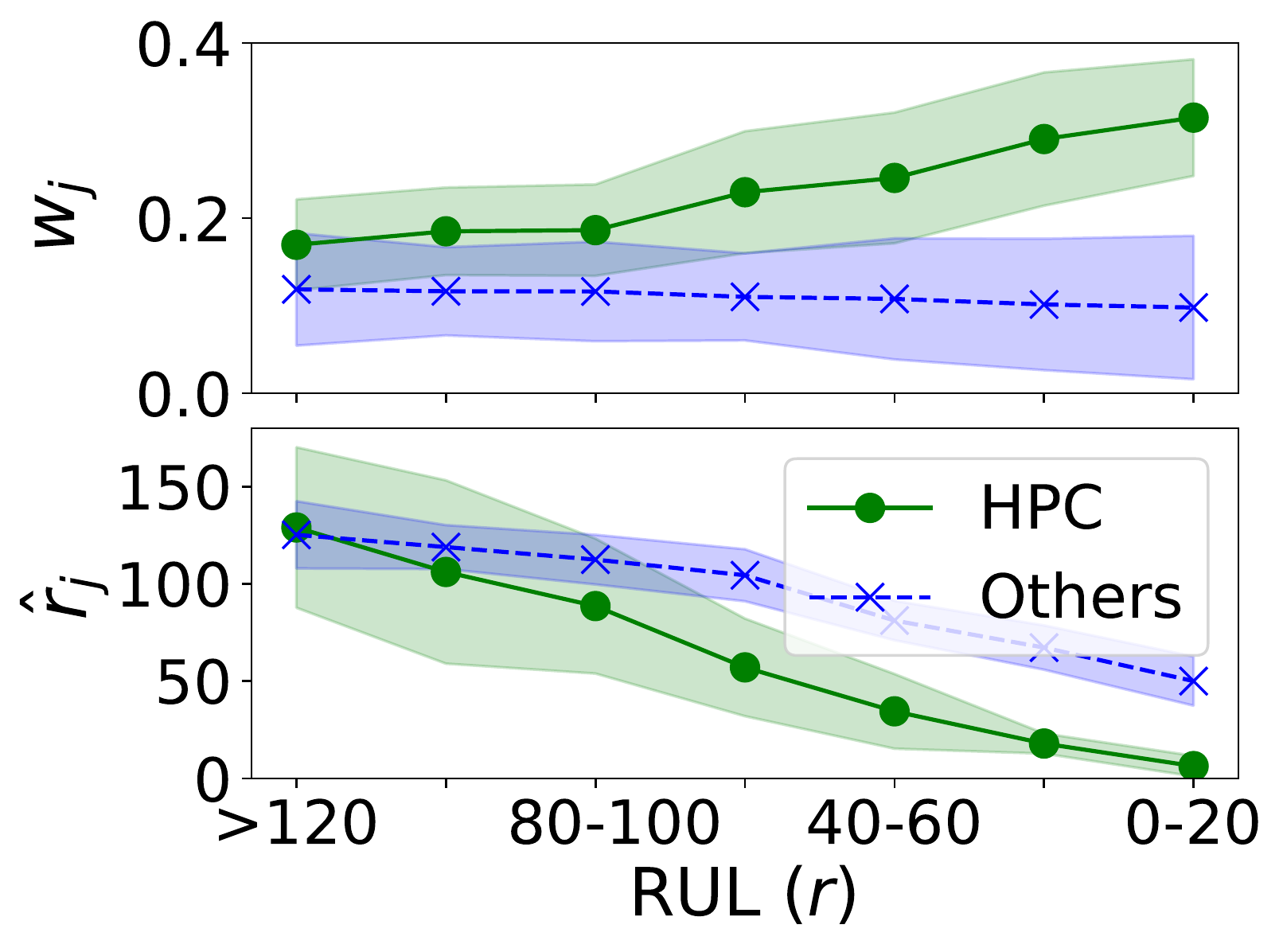}
		\caption{FD002}
	\end{subfigure}
	\caption{Average attention weight and RUL estimates w.r.t ground truth RUL for the HPC node with impending failure versus other nodes. \label{fig:attention}}
\end{figure}

	\noindent \textbf{(3)} \textit{Preliminary Analysis of Attention Mechanism (refer Eq. \ref{eq:attn})}: 
	For FD001 and FD002 datasets, the faults in all instances are known to originate in the HPC module\footnote{For FD003 and FD004, the ground truth information about the faulty node is not available, hence not analyzed.} (refer Fig. \ref{fig:structure-used}). 
	It is therefore expected that as the degradation increases, the behavior of sensors associated with the HPC module will depict signatures for detecting the impending failure, and in turn, estimating the RUL. 
	We observe that GNMR implicitly tends to capture and leverage this behavior in some cases: we analyze the attention weights $w_j$s and the corresponding contribution $\hat{r}_j$ to the RUL estimate from the HPC node versus the remaining nodes.
	As shown in Fig. \ref{fig:attention}(b), we observe that for FD002 dataset, the attention for the faulty node (HPC module) increases as the target RUL decreases (i.e. as the engines approach failure), while the attention tends to decrease for the remaining nodes. 
	However, this trend is not observed in FD001 Fig. \ref{fig:attention}(a).
	In future, it will be interesting to see if this can be explicitly ensured: while other modules may provide additional information to track the health degradation of a particular module, it may be useful to bias the attention to the module-of-interest.

\section{Conclusion and Future Work\label{sec:discussion}}
We have proposed an approach to incorporate the readily available information about the modularized structure of complex equipment into deep learning models via gated graph neural networks (GNNs).
To the best of our knowledge, our work provides a first set of results for leveraging GNNs in the increasingly important area of automated equipment health monitoring. 
We analyze the heavily-benchmarked aircraft turbofan engine dataset through the lens of structure-aware deep learning, potentially bridging the gap between deep learning and the domain-knowledge aware approaches. We hope that this work inspires future research in leveraging equipment structure to model the behavior of complex systems (such as power plants) with interesting applications like optimization and anomaly detection.  
While the graph structure is readily available in most practical applications as part of domain knowledge, it may not be optimal in terms of reflecting the dependencies between sensors at a node or between sensors across nodes. It will be interesting to explore if the optimal graph structure can itself be learned starting from the domain knowledge-based initial graph structure.

\bibliographystyle{aaai}
\bibliography{bibTeX/aaai,bibTeX/dise}

\section*{Datasets and Pre-processing Details}

The datasets contain time series of readings for 24 sensors (21 sensors and 3 operating condition variables), such that each cycle in the life of an engine provides a 24-dimensional vector.
The sensor readings in the training set are available from beginning of usage of the engine till the end of life, while those in the test split are clipped at a random time prior to the failure such that the instances are operational at the last available cycle, and the goal is to estimate the RUL for these test instances. 
Refer Table \ref{tab:datastats} for details of the datasets.

\begin{table}[h]
	\caption{Basic statistics of the four datasets used. \label{tab:datastats}}
	\centering
	\scalebox{0.7}{
		\begin{tabular}{|l|c|c|c|c|}\hline
			{Dataset $\rightarrow$} & {\bfseries FD001} & {\bfseries FD002} & {\bfseries FD003} & {\bfseries FD004} \\\hline
			Instances (training set)& 80 & 208 & 80 & 199 \\
			Instances (validation set) & 20 & 52 & 20 & 50\\
			Instances (test set) & 100 & 259 & 100 & 248\\
			Operating conditions & 1 & 6 & 1 & 6 \\
			Fault Modes & 1 & 1 & 2 &2 \\	\hline		
		\end{tabular}
	}
\end{table}

As commonly used in RUL estimation approaches \cite{zheng2017long,babu2016deep,li2018remaining}, we also consider an upper bound $r_u$ on the possible values of target RUL during training as, in practice, it is not possible to predict too far ahead in future. So if $r > r_u$, we clip the value of $r$ to $r_u=130$.
The targets are then normalized to $\frac{r}{r_u}$, such that the targets in training are in the range 0-1.
We use min-max normalization technique to normalize input time series sensor wise using the minimum and maximum value of each sensor (in the range of -1 to 1) from the training set.
The time series for each engine instance is then divided into windows of length $T=100$ with window-shift of 5, refer Table \ref{tab:datastats_info} for details of the number of resulting time series after windowing.
We use suitable pre-padding with mean value of sensor readings to ensure same length for each time series for both GRU-MR and GGNN.
\begin{table}[h]
	\caption{Number of time series windows. \label{tab:datastats_info}}
	\centering
	\scalebox{0.7}{
		\begin{tabular}{|l|c|c|c|c|}\hline
			{Dataset $\rightarrow$} & {\bfseries FD001} & {\bfseries FD002} & {\bfseries FD003} & {\bfseries FD004} \\\hline
			Training& 1,875 & 4,688 & 2,129 & 5,383 \\
			Validation & 411 & 1,287 & 533 & 1,451\\
			Test & 100 & 259 & 100 & 248\\\hline
		\end{tabular}
	}
\end{table}

\end{document}